\documentclass[conference]{IEEEtran}
\IEEEoverridecommandlockouts
\usepackage{cite}
\usepackage{amsmath,amssymb,amsfonts}
\usepackage{algorithmic}
\usepackage{graphicx}
\graphicspath{{figures/}}
\usepackage{textcomp}
\usepackage{xcolor}

\usepackage{url}

\def\BibTeX{{\rm B\kern-.05em{\sc i\kern-.025em b}\kern-.08em
    T\kern-.1667em\lower.7ex\hbox{E}\kern-.125emX}}
\begin{document}

\title{Misaka: Interactive Swarm Testbed for Smart Grid Distributed Algorithm Test and Evaluation\\
\thanks{This work was supported in part by National Natural Science Foundation of China (No. 51777102), in part by National Training Program of Innovation and Entrepreneurship for Undergraduates. The authors wish to express their deep appreciation and gratitude to Professor Haipeng Mi, Yu Peng, Hua Tong and other colleagues in Future Lab, Tsinghua University for their guidance. The authors are also very grateful to Yihan Jia for her modification of this paper.
}
}

\author{\IEEEauthorblockN{Tingliang Zhang}
\IEEEauthorblockA{\textit{Dept. Electrical Engineering} \\
\textit{Tsinghua University}\\
Beijing, China \\
zhangtl16@mails.tsinghua.edu.cn}
\and
\IEEEauthorblockN{Haiwang Zhong}
\IEEEauthorblockA{\textit{Dept. Electrical Engineering} \\
\textit{Tsinghua University}\\
Beijing, China \\
zhonghw@tsinghua.edu.cn}
\and
\IEEEauthorblockN{Zhenfei Tan}
\IEEEauthorblockA{\textit{Dept. Electrical Engineering} \\
\textit{Tsinghua University}\\
Beijing, China \\
}
\and
\IEEEauthorblockN{Xinfei Yan}
\IEEEauthorblockA{\textit{Dept. Electrical Engineering} \\
\textit{Tsinghua University}\\
Beijing, China \\
}
}

\maketitle

\begin{abstract}
    In this paper, we present Misaka, a visualized swarm testbed for smart grid algorithm evaluation, also an extendable open-source open-hardware platform for developing tabletop tangible swarm interfaces. The platform consists of a collection of custom-designed 3 omni-directional wheels robots each 10 cm in diameter, high accuracy localization through a microdot pattern overlaid on top of the activity sheets, and a software framework for application development and control, while remaining affordable (per unit cost about 30 USD at the prototype stage). We illustrate the potential of tabletop swarm user interfaces through a set of smart grid algorithm application scenarios developed with Misaka.
\end{abstract}


\section{Introduction}

Most recently, an increasing penetration of distributed energy resources (DERs) in Energy Internet imposes great challenges on conventional centralized economic dispatch\cite{yan2019consensus}.

The centralized controller requires a high-bandwidth communication infrastructure and a high level of connectivity, poses reliability concerns, and is prone to modeling errors. Moreover, both the future power grid and the communication network are likely to have a variable topology, which further undermines the efficacy of centralized mechanisms\cite{binetti2013distributed}.

In the smart grid environment, the decision-making procedures are moving from centralized to distributed frameworks\cite{yang2011communication}.

But distributed algorithms are relatively difficult to understand, and need communication test to determine feasibility in a real hardware scenario. So we present a swarm testbed to visualize and evaluate the algorithm and communication design of those newly proposed distributed frameworks for smart grid. A Misaka is a hardware and software system: a small omni-wheel robot with position sensing capabilities that can be freely arranged and repositioned on any horizontal surface, both through user manipulation and computer control. 

Due to Misakas’ ability to quickly and freely reconfigure themselves spatially, a collection of Misakas can act as a display and can provide meaningful user output such as multiple scatterplots and line charts. Due to their ability to sense user actions, Misakas can also support rich input. For example, users can move or rotate Misakas to change parameter and manipulate others Misakas.The system is relatively lightweight and only requires any surface, such as a sheet of paper or a game board, with simple printed structured patterns for real-time high-accuracy 2D localization\cite{yu2019swarm}.

To stimulate future research on swarm user interfaces, we distribute our Misaka tabletop swarm user interface platform in open-source and open-hardware.

In summary, our contributions are:

\begin{itemize}
    \item The first open-source hardware/software platform for smart grid algorithm evaluation with tabletop swarm user interfaces
    \item Redefinition for algorithm visualization with several implemented examples
    \item A set of scenarios to illustrate the unprecedented possibilities offered by Misaka swarm and by tabletop swarm user interfaces in general
    \item A common platform for any algorithm visualization, and other interactive swarm user interfaces
\end{itemize}

Furthermore, as benefits, Misaka:

\begin{itemize}
    \item are modular, can be extended with powerful platform, such as NVIDIA Jetson NANO.
    \item can simulate real smart grid communication scenarios with Zigbee modular on board
    \item can interact with user moving and turning the robot or using the illuminated capacitive touch keys on the robot 
    \item can be manipulated either individually or collectively
    \item are small enough to also act as “pixels” of a physical display
    \item can coexist in large numbers
    \item are lightweight, can operate on any horizontal surface, and relatively cost-effective: about 30 USD each now, down to \$15 if mass manufactured.
\end{itemize}

\section{Consensus Algorithm Design}
\label{sec:Algorithm}

To explain Misaka usage in smart grid algorithm evaluation scenario, we propose a simplified average consensus algorithm to solve some distributed economic dispatch (DED) problems based on a consensus-based information exchange architecture, in which a generator only communicates with its neighbors. The algorithm is a linear consensus protocol based on Markov chain and incremental cost consensus (ICC) method.

Considering the actual situation, most complex algorithms require a central coordination node to distribute the transition matrix. The algorithm we propose does not depend on the central node (coordinator) at the hardware level, aka completely decentralized. In addition, the operation of each node is very simple, and it has been proved that this algorithm converges in a strongly connected system.

A distributed solution must be supported by a communication network that provides its information flow, i.e., each generating unit exchanges information with a subset of other units to make autonomous decisions\cite{binetti2013distributed}. The communication connections can be described by a connected graph $G=(V, E)$ where $V=\left\{\nu_{1}, \nu_{2}, \ldots, \nu_{n}\right\}$ is the set of nodes, and $E \subset V \times V$
is the set of edges. An edge $\left(\nu_{j}, \nu_{i}\right) \in E$ from node $j$ to node $i$ exists if node $i$ can receive information from node $j$.

We use the adjacency matrix A of the directed graph to describe its communication topology:

\begin{equation}
    \mathbf{A}=\left[a_{i j}\right]_{N \times N}
\end{equation}

\begin{equation}
    a_{\mathrm{ij}}=\left\{\begin{array}{ll}
    {1} & {\text { if } j \text{ can receive information directly from } i } \\
    {0} & {\text { Otherwise }}
    \end{array}\right.
\end{equation}

The information flow can be described by a weighted graph, whose transition matrix $Q$ are composed by communication weights:

\begin{equation}
    q_{i j}=\frac{a_{i j}}{\sum_{k=1}^{N} a_{i k}}
\end{equation}

Then the consensus protocol can be formulated as follows.

\begin{equation}\boldsymbol{s}^{(k+1)}=\boldsymbol{Q} \boldsymbol{s}^{(k)}\end{equation}

The physical meaning of the algorithm is: at the same time during each iteration, all weights connected to the node are averaged and assigned to this node.

In terms of the transition matrix: each element in the $Q$ matrix is the corresponding adjacent matrix element divided by the node outdegree (the total number of edges emitted).

\begin{figure}[htbp] 
    \centering
    \includegraphics[width=0.6\columnwidth]{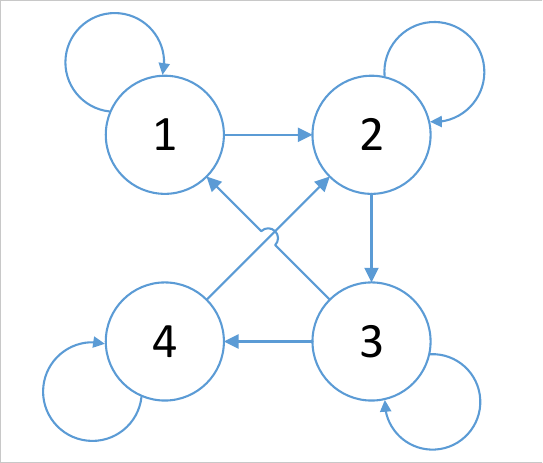}
    \caption{4-node directed graph}
    \label{fig:Directed-graph}
\end{figure}

To demonstrate the algorithm performance, we propose two typical cases:

Case 1 is a 4-generator directed graph given in Fig~\ref{fig:Directed-graph}.

The adjacency matrix A is:

\begin{equation}
    A=\left[\begin{array}{cccc}
    {1} & {1} & {0} & {0} \\
    {0} & {1} & {1} & {0} \\
    {1} & {0} & {1} & {1} \\
    {0} & {1} & {0} & {1}
    \end{array}\right]
\end{equation}

The transition matrix is:

\begin{equation}
    Q=\left[\begin{array}{cccc}
    {\frac{1}{2}} & {\frac{1}{2}} & {0} & {0} \\
    {0} & {\frac{1}{2}} & {\frac{1}{2}} & {0} \\
    {\frac{1}{3}} & {0} & {\frac{1}{3}} & {\frac{1}{3}} \\
    {0} & {\frac{1}{2}} & {0} & {\frac{1}{2}}
    \end{array}\right]
\end{equation}

\begin{figure}[htbp]
    \centering
    \includegraphics[width=\columnwidth]{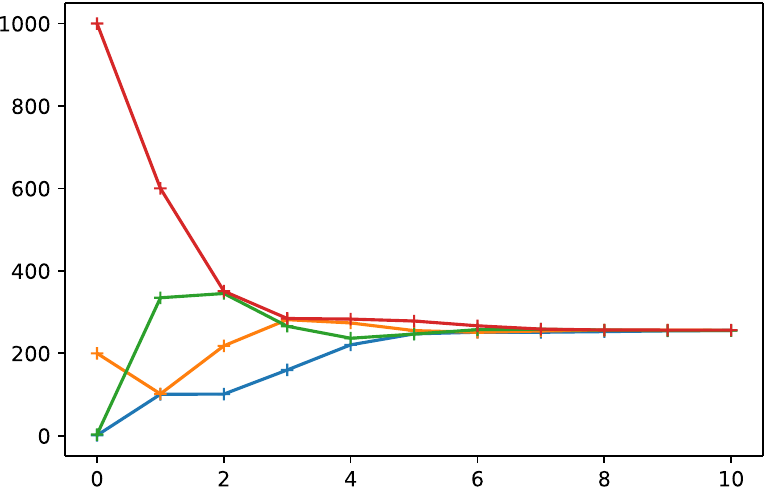}
    \caption{Case 1 price curve of each unit}
    \label{fig:Result-1234}
\end{figure}

Iterate 10 times with the initial values 1, 2, 3, 4. With the iteration, the price curve of each unit is shown in Fig~\ref{fig:Result-1234}.

As the iterations approach consensus data are shown in Table~\ref{tab:Result-1234}. In $X_{i,j}$, $i$ is number of iterations and $j$ is node number.


\begin{table}[htbp]
    \centering
    \caption{Case 1 iteration data}
    \begin{tabular}{lllll}
    \hline
    \hline
    $X_{i,j}$   & 0        & 1        & 2        & 3        \\ \hline
    0  & 1        & 2        & 3        & 4        \\ 
    1  & 1.5      & 2.5      & 2.666667 & 3        \\ 
    2  & 2        & 2.583333 & 2.388889 & 2.75     \\ 
    3  & 2.291667 & 2.486111 & 2.37963  & 2.666667 \\ 
    4  & 2.388889 & 2.43287  & 2.445988 & 2.576389 \\ 
    5  & 2.41088  & 2.439429 & 2.470422 & 2.50463  \\ 
    6  & 2.425154 & 2.454925 & 2.461977 & 2.472029 \\ 
    7  & 2.44004  & 2.458451 & 2.453054 & 2.463477 \\ 
    8  & 2.449246 & 2.455752 & 2.45219  & 2.460964 \\ 
    9  & 2.452499 & 2.453971 & 2.454133 & 2.458358 \\ 
    10 & 2.453235 & 2.454052 & 2.454997 & 2.456165 \\ 
    \hline
    \hline
    \end{tabular}
    \label{tab:Result-1234}
\end{table}

For a non-strongly connected graph, the strongly connected component of the directed graph will produce a global error iteration result because of the unidirectionally connected nodes. It is necessary to ensure that the communication error nodes are eliminated at all times. The handshake protocol can be used to ensure the reliability of two-way communication.

As an example, Case 2 is a 10-node system, the values of nodes 1-10 correspond to 1-10 respectively, and the original non-strongly connected graph adjacency matrix A is :

\begin{equation}
    A=\left(\begin{array}{cccccccccc}
    1 & 1 & 0 & 0 & 1 & 1 & 0 & 1 & 0 & 0 \\
    0 & 1 & 1 & 0 & 0 & 0 & 1 & 0 & 1 & 0 \\
    1 & 0 & 1 & 1 & 0 & 0 & 1 & 1 & 0 & 1 \\
    0 & 1 & 0 & 1 & 0 & 1 & 0 & 0 & 1 & 0 \\
    0 & 0 & 0 & 0 & 1 & 0 & 0 & 0 & 0 & 1 \\
    0 & 0 & 0 & 0 & 0 & 1 & 0 & 1 & 0 & 0 \\
    0 & 0 & 0 & 0 & 0 & 0 & 1 & 0 & 0 & 1 \\
    0 & 1 & 0 & 0 & 1 & 1 & 0 & 0 & 0 & 0 \\
    0 & 0 & 0 & 0 & 0 & 0 & 0 & 0 & 1 & 0 \\
    0 & 0 & 0 & 0 & 1 & 0 & 1 & 0 & 1 & 1
    \end{array}\right)
\end{equation}

\begin{figure}[htbp]
    \centering
    \includegraphics[width=\columnwidth]{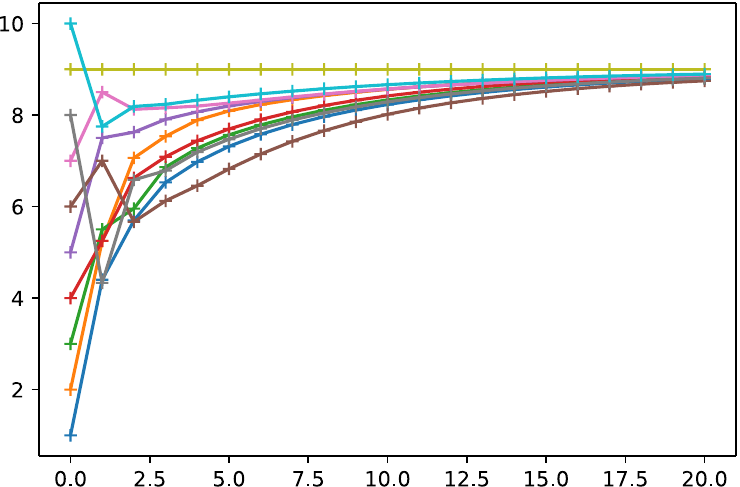}
    \caption{Non-strongly connected graph iteration curve}
    \label{fig:123456-Error}
\end{figure}

Note that node 9 has no way to receive information from other nodes and becomes a strongly connected component separated from other nodes, make the entire network erroneously converge to the value of node 9, which is shown in Fig~\ref{fig:123456-Error}.

\begin{figure}[htbp]
    \centering
    \includegraphics[width=\columnwidth]{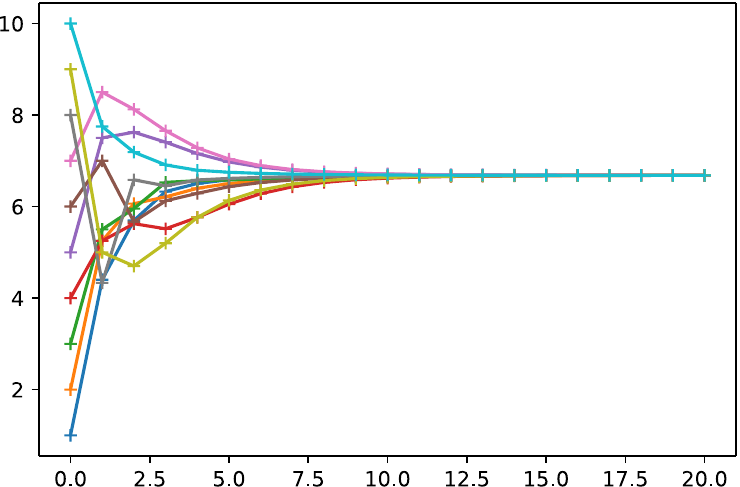}
    \caption{Case 2 strongly connected graph iteration curve}
    \label{fig:123456-Correct}
\end{figure}

After connecting the directed edges from nodes 9 to 1, it will become a strongly connected system and converge normally, which is shown in Fig~\ref{fig:123456-Correct}.

\section{UI design with Misaka}

To illustrate the potential of tabletop swarm user interfaces, we present a set of smart grid algorithm development application scenarios developed with Misaka.

\subsection{Dynamic iteration process physicalization}

\begin{figure}[htbp]
    \centering
    \includegraphics[width=\columnwidth]{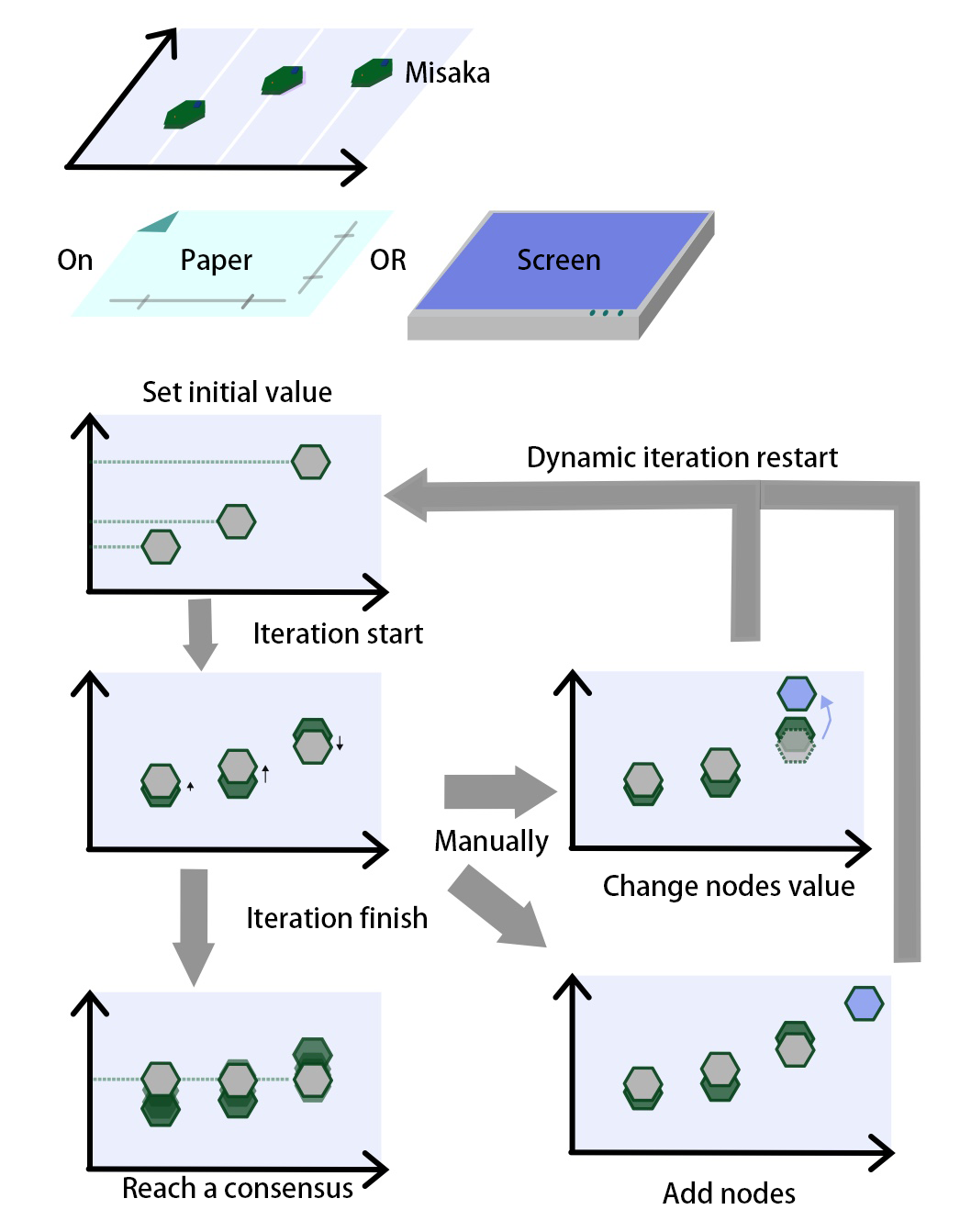}
    \caption{Interactive dynamic iteration}
    \label{fig:iteration}
\end{figure}

In the dynamic iterative visualization mode, the working surface is regarded as a two-dimensional rectangular coordinates. Each Misaka represents a generator node, and the vertical axis position of Misaka in the rectangular coordinates represents its current output, changing with each iteration. This system also simulate the communication process and perform decentralized autonomous optimization. When the number of nodes (newly added or withdrawn) or any node data changes, the system will restart dynamic iteration. The initial value can be set by the position of Misaka at the beginning, also Misaka can be manually moved at any time during or after the iteration to change the value of nodes. The sketch map of dynamic iterative visualization is shown in Fig~\ref{fig:iteration}. 

In other words, we can interact with the algorithm, more interaction methods will be further discussed in the next section.

The relationship between physical world and algorithms is shown as Table~\ref{tab:Real-Unreal}.

\begin{table}[htbp]
    \centering
    \caption{Relationship between physical world and algorithms}
    \begin{tabular}{@{}ll@{}}
    \hline
    \hline
    physical world          & algorithms                               \\ \hline
    working surface         & two-dimensional rectangular coordinates  \\
    move forward            & vertical axis                            \\
    a Misaka                & a node                                   \\
    place new Misaka        & add new node and set initial value       \\
    remove a Misaka         & disconnect a node                        \\ 
    \hline
    \hline
    \end{tabular}
    \label{tab:Real-Unreal}
\end{table}


\subsection{Information transmission visualization}


Misakas with LCD display above can serve as a communication procedures visualization platform. Some Misakas representing nodes show current parameter values and calculate process on the LCD screen. Others transmit data on edges as messengers, whose screen presents carried packages.

\begin{figure}[htbp]
    \centering
    \includegraphics[width=\columnwidth]{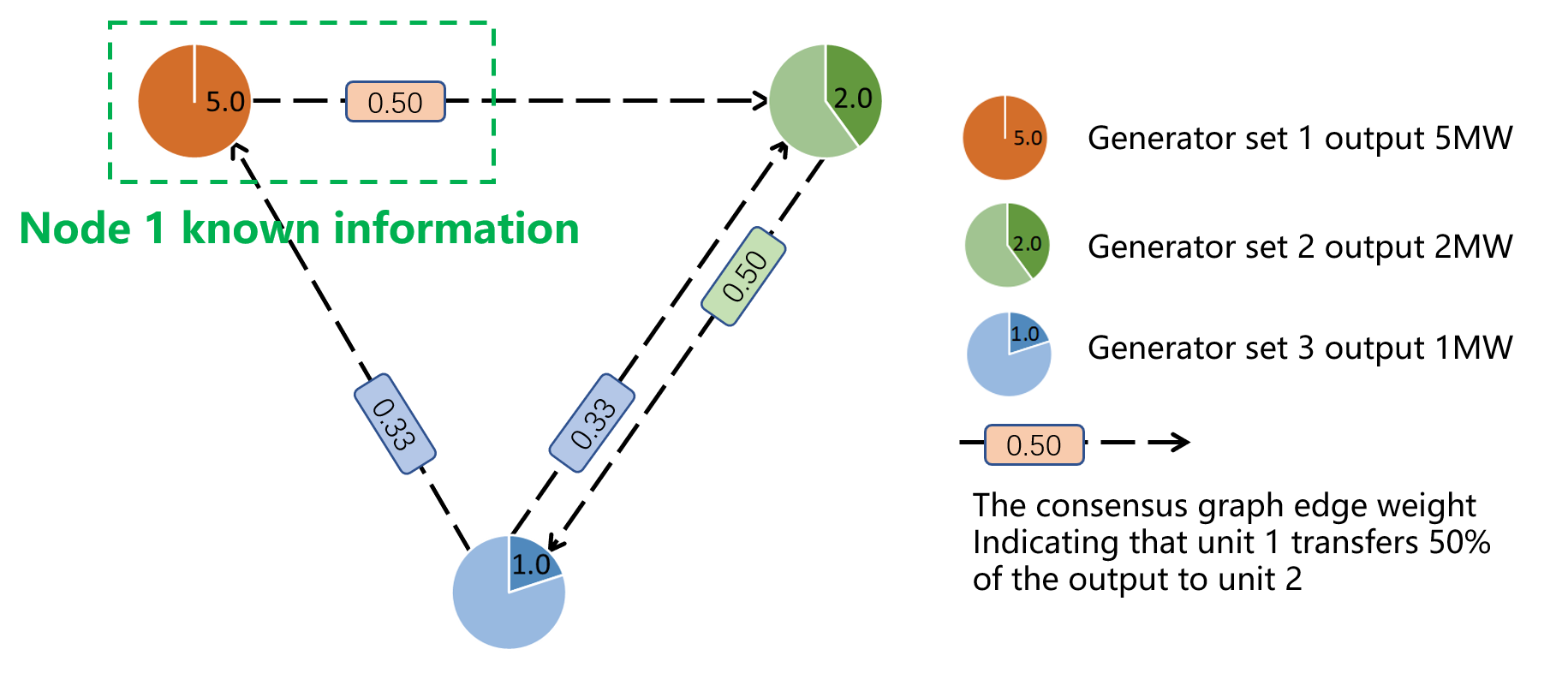}
    \caption{3-node information transmission visualization system}
    \label{fig:PPT0}
\end{figure}

To illuminate information transmission visualization with Misaka, we assume there is a 3-node system, which is shown in Fig~\ref{fig:PPT0}. The initial output is random, assuming 5MW, 2MW, and 1MW. And the transition matrix Q is:

\begin{equation}
    Q=\left[\begin{array}{ccc}
    0.50 & 0 & 0.33 \\
    0.50 & 0.50 & 0.33 \\
    0 & 0.50 & 0.33
    \end{array}\right]
\end{equation}

According to the principle of equal incremental rate, it is necessary to make the output power of nodes become 1: 1: 1 while keeping the total output unchanged where economy is optimal.

In this example, three fixed Misakas represent nodes, and one Misaka on each directional edge represents a data packet transmitted by the communication protocol.

\begin{figure*}[ht]
    \centering
    \includegraphics[width=5cm]{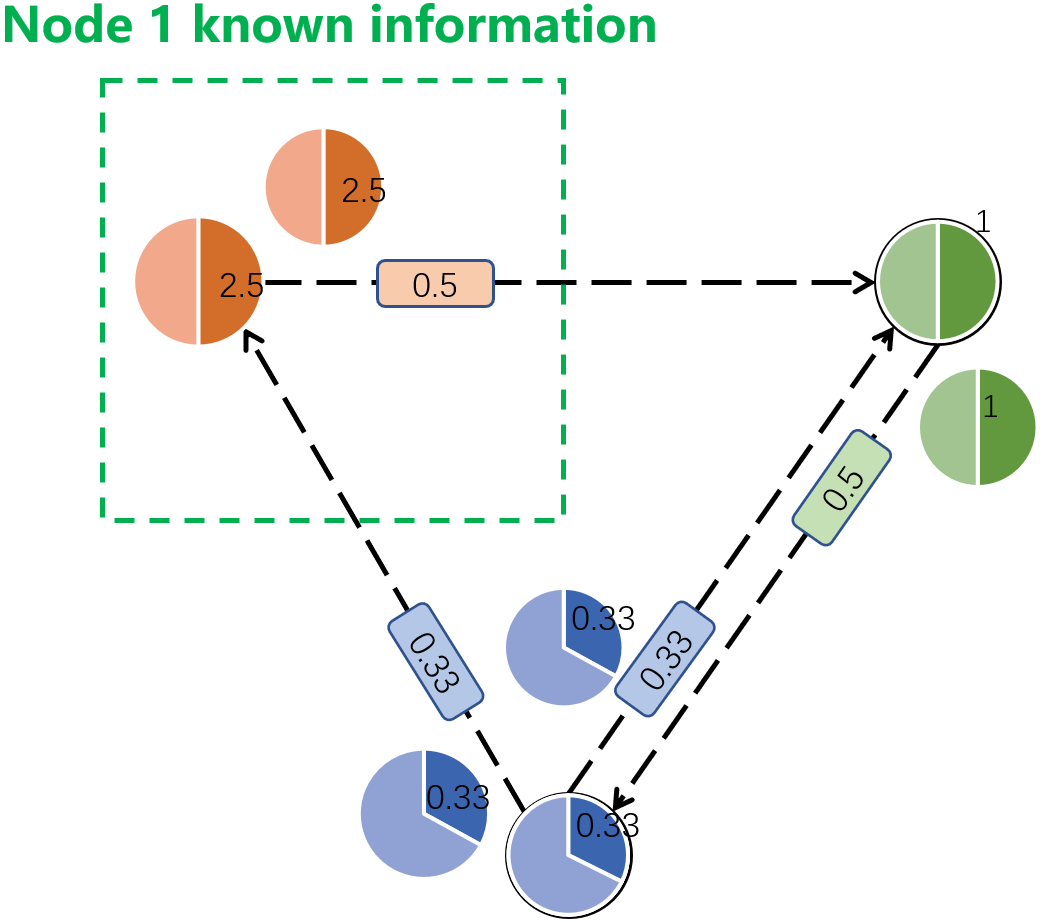}
    \includegraphics[width=5cm]{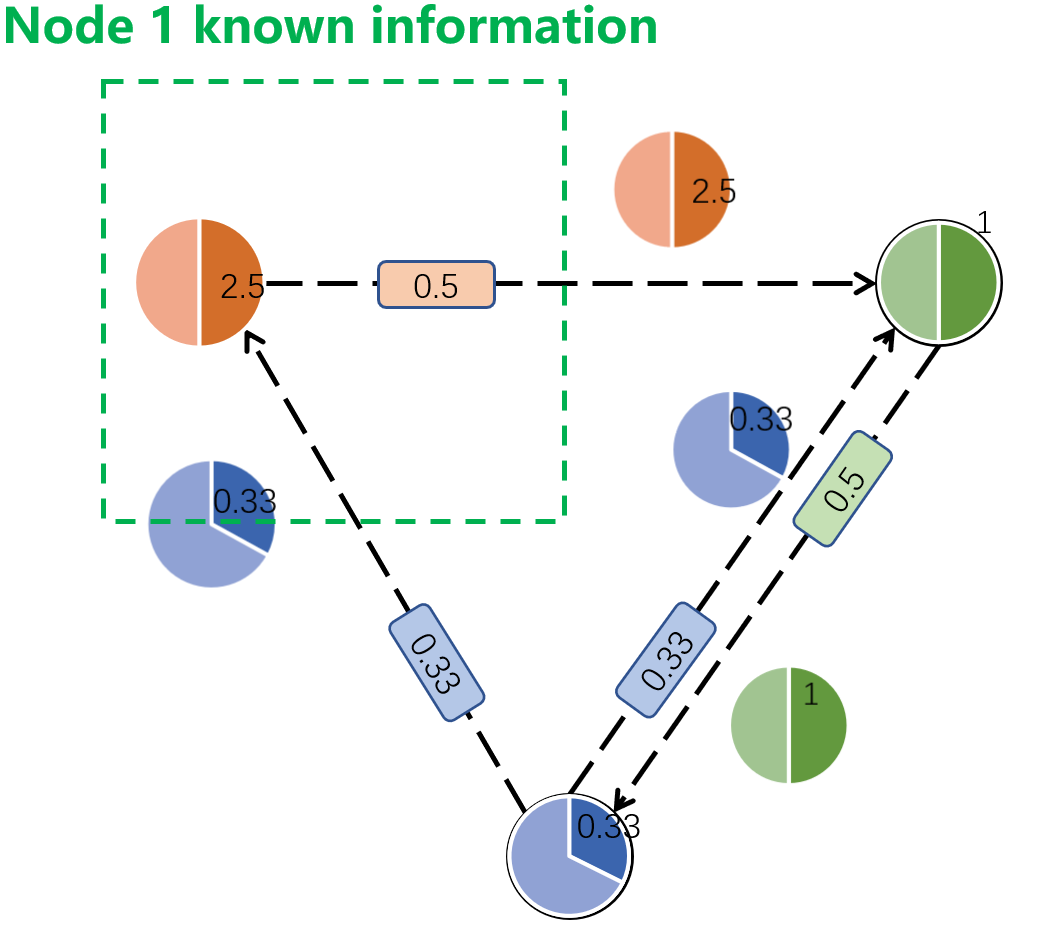}
    \includegraphics[width=5cm]{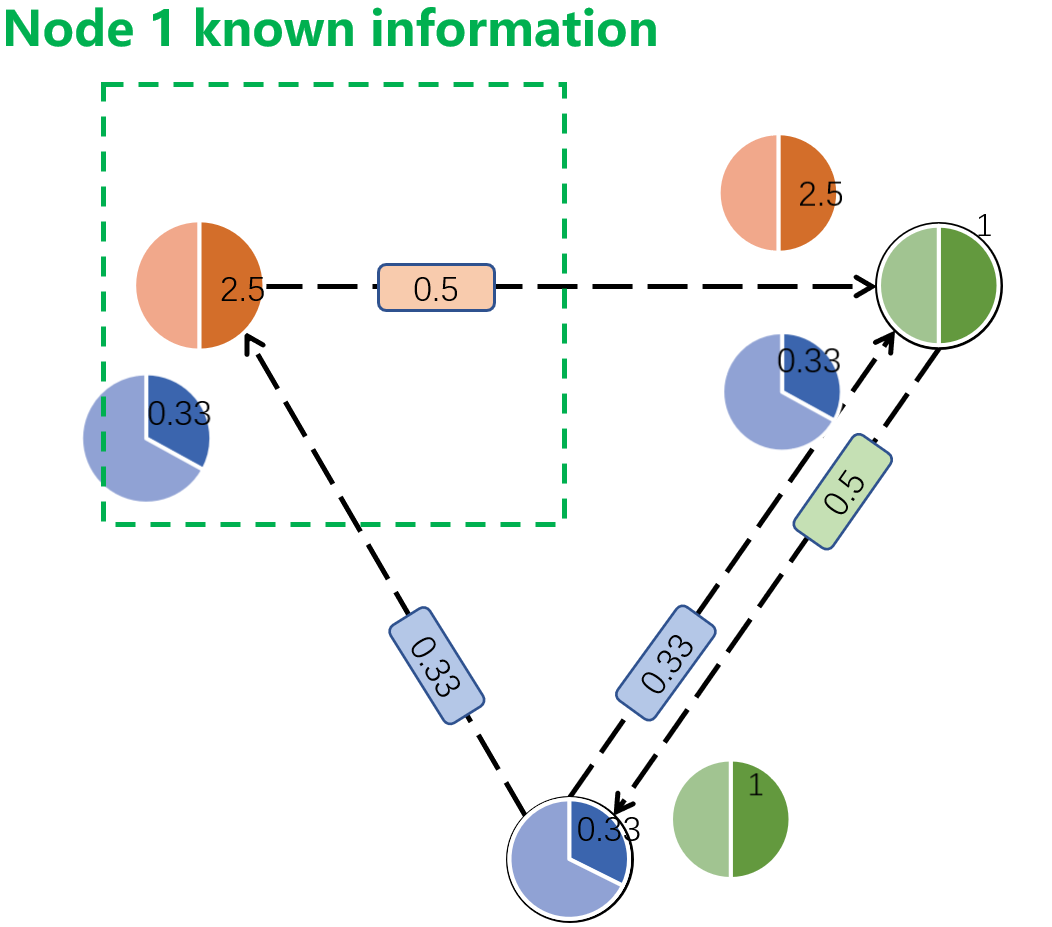}
    \caption{Information transmission visualization}
\label{fig:PPT}
\end{figure*}

When any data packet is sent, the corresponding Misaka of the directional edge will move to the sending node (within a certain distance that the two can communicate with each other) to receive the data packet and display it on its screen. After that, it will move towards the target node. After reaching a certain range around the target communication node, it will send the data packet to the target node, which will perform a series of operations and display the process on its screen. The process is shown in Fig~\ref{fig:PPT}.

\subsection{Time-series navigation}


\begin{figure}
    \begin{minipage}{0.49\columnwidth}
      \centering
      \includegraphics[width=0.99\columnwidth]{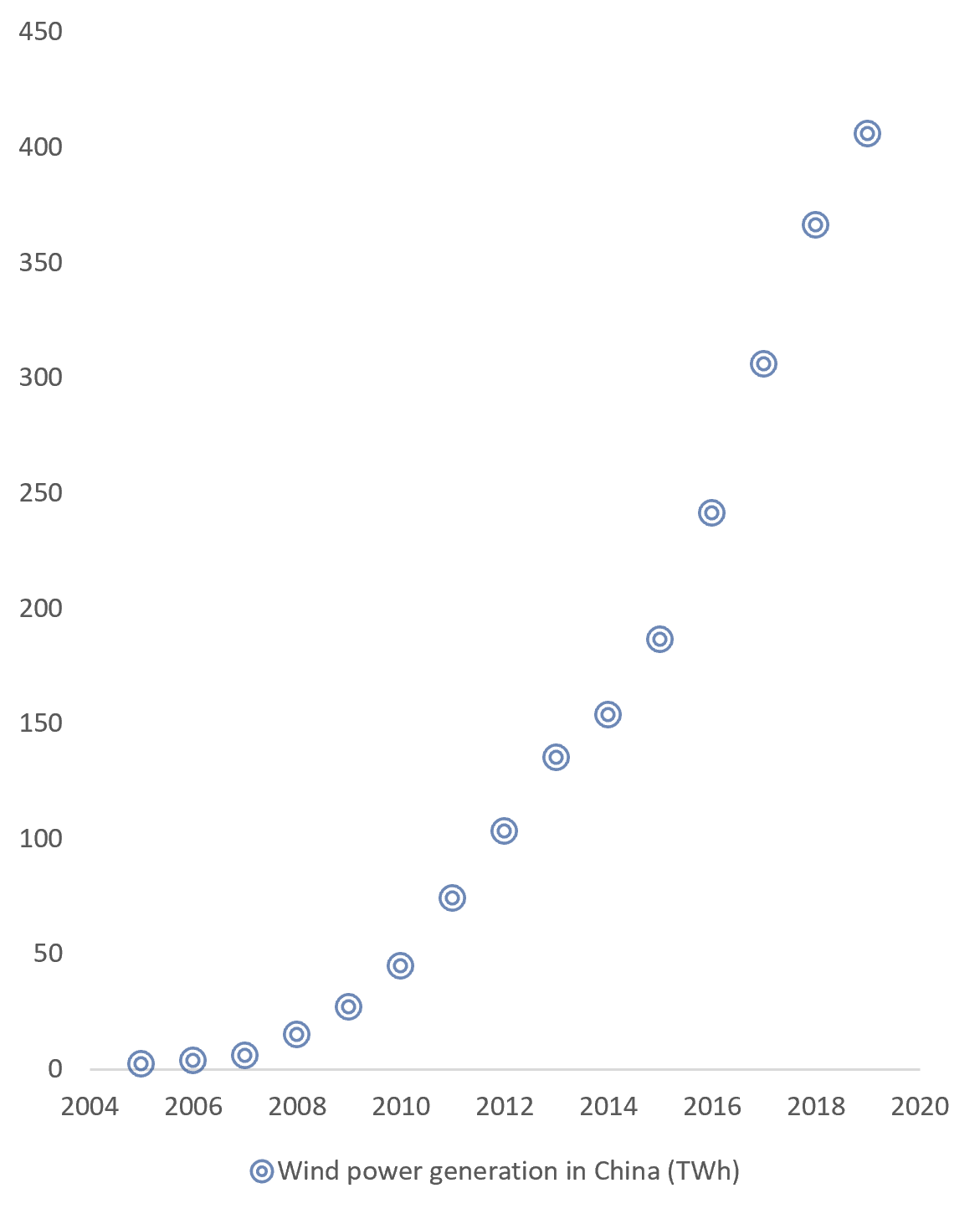}
      \caption{Single scatterplots visualization with Misakas}
      \label{fig:Single}
    \end{minipage}\hfill
    \begin{minipage}{0.49\columnwidth}
      \centering
      \includegraphics[width=0.99\columnwidth]{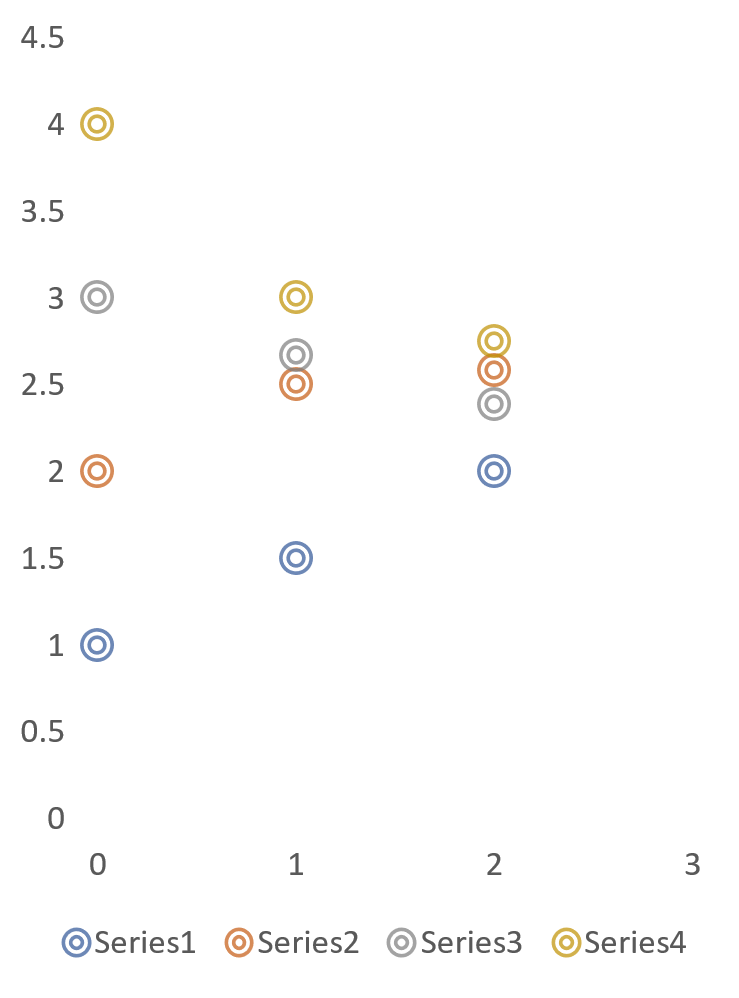}
      \caption{Multiple scatterplots}
      \label{fig:Multiple}
    \end{minipage}
\end{figure}

We use Misaka to visualize and navigate in time-series data, Wind power generation in China from 2005 to 2019 for example. In Fig~\ref{fig:Single}, each circle represent a Misaka. Beside this plot, there are two other Misakas act as widgets to let users customize the display and navigate the data. The two Misakas specify the time range, they act like a range slider\cite{ahlberg1992dynamic}. For instance, moving those two Misakas apart, the line chart will show a wider time range.


\subsection{Multiple scatterplots}

Scatterplots is a type of plot using Cartesian coordinates to display values for typically two variables for a set of data. 


Misakas with RGB LED on top can use color to represent different sets of data, or another scalar variable.

We use Case 1 in Section~\ref{sec:Algorithm} to demonstrate this. As Fig~\ref{fig:Multiple} shows, Misakas’ LED color represents different data series, and the position of each Misaka tells two scalar variables. 

Another example is placing Misakas on any map(or screen which display map), the location of Misaka means the correspond latitude and longitude on the map. The color of RGB LED represents scalar variable value.

\begin{figure}[htbp]
    \centering
    \includegraphics[width=\columnwidth]{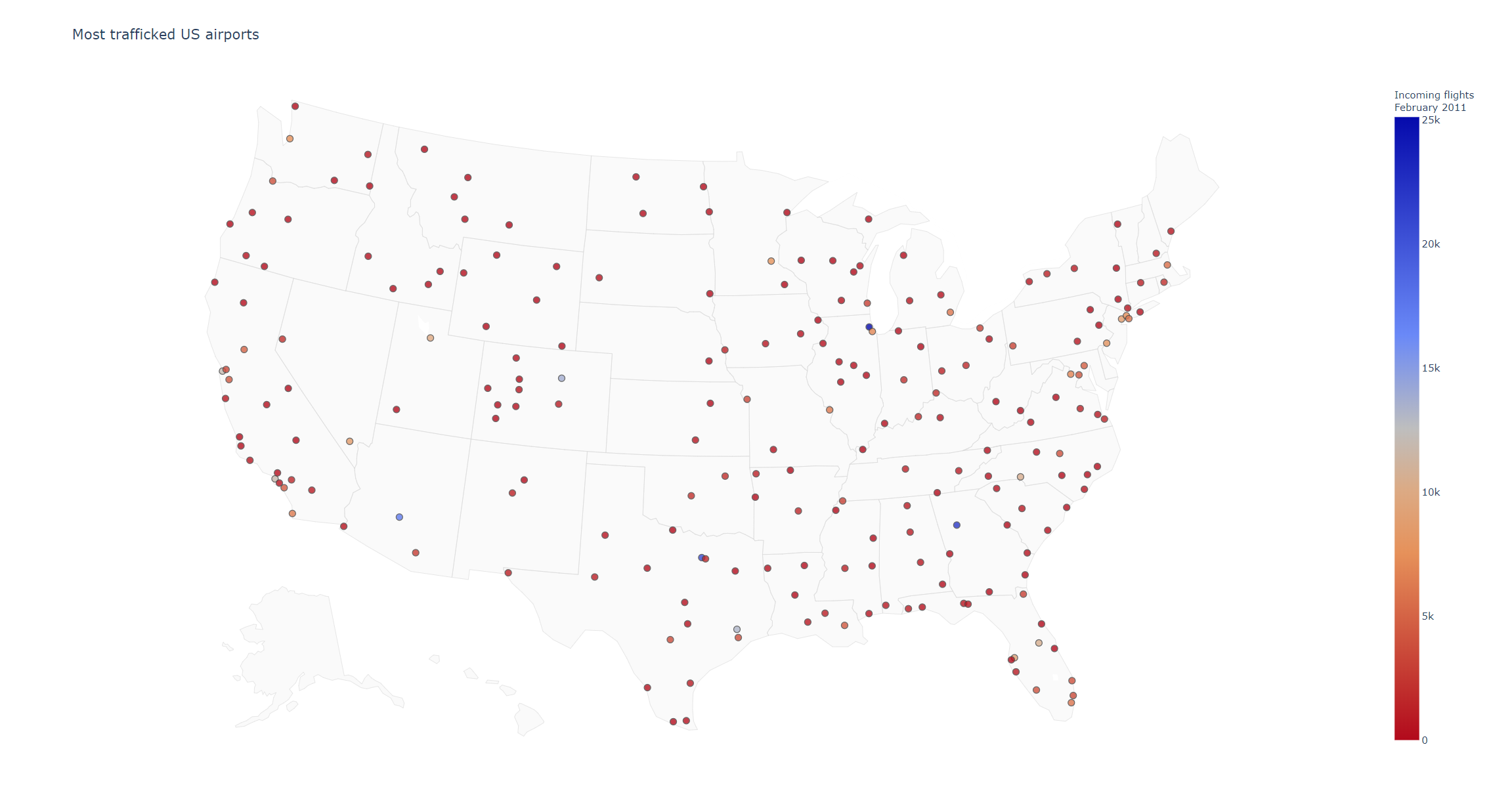}
    \caption{Most trafficked US airports visualization with Misakas}
    \label{fig:scatterplots2}
\end{figure}

We use Misaka on U.S. map to visualize US airport traffic(incoming flights on February 2011) of most trafficked US airports. Shown in Fig~\ref{fig:scatterplots2}.

\section{Hardware design}

There are many tangible robots such as Zooids\cite{le2016zooids} and Cellulo\cite{ozgur2017cellulo}. We refer to their design and carefully designed Misaka testbed according to our specific application scenario.

\begin{figure}[htbp]
    \centering
    \includegraphics[width=0.6\columnwidth]{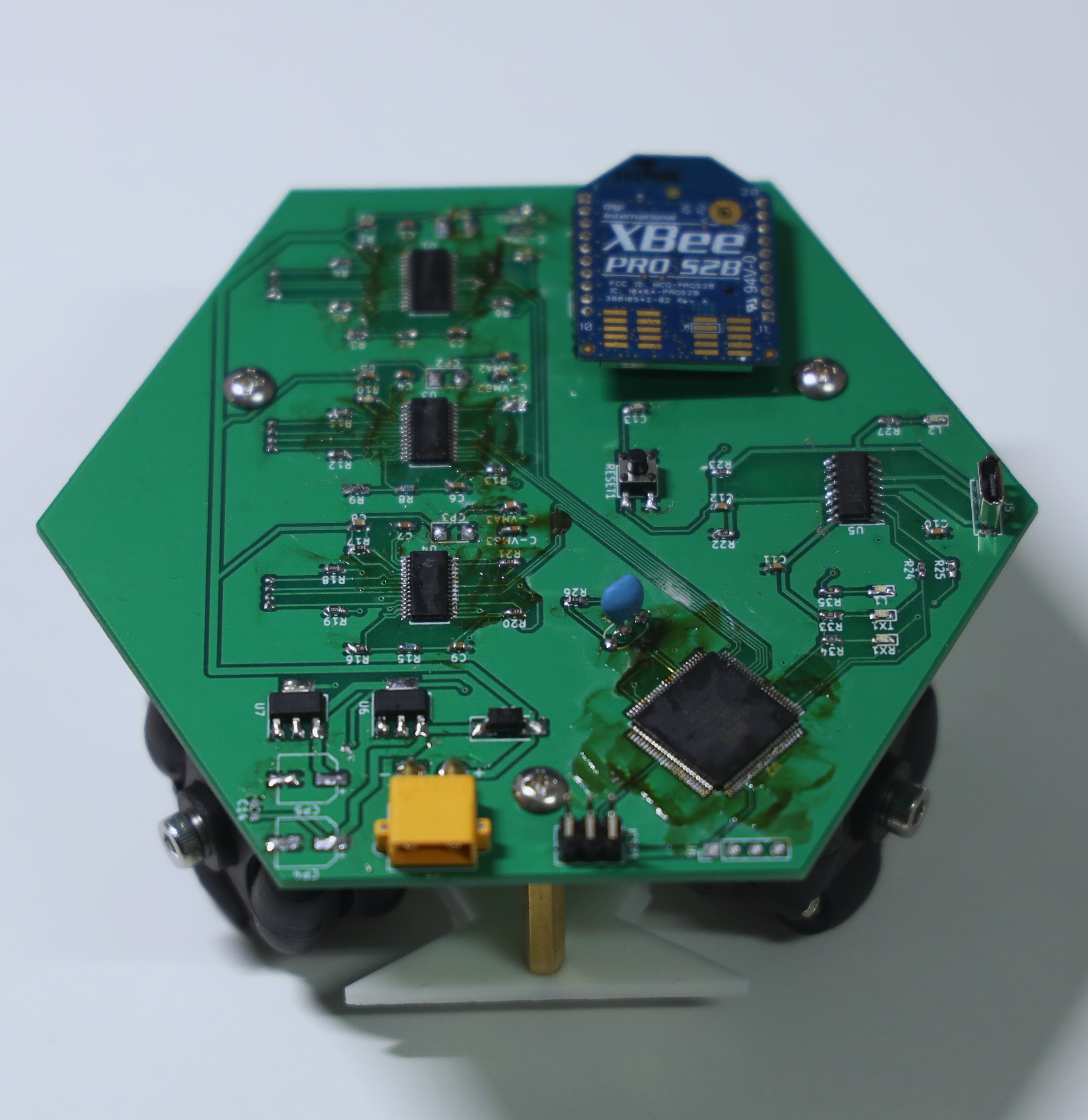}
    \caption{Misaka core model}
    \label{fig:model}
\end{figure}

Misakas are small custom-made robots as shown in Fig~\ref{fig:model}, their dimensions are 100 mm in diameter and 50 mm in height. Each robot is powered by a 450mAh 2S 7.4V LiPo battery and uses micro stepper motor driven wheels.

Traditional two wheels differential drive normally used on mobile robots have manoeuvrability limitations and take time to sort out\cite{ribeiro2004three}. 

An Omni wheel has an axis perpendicular to the axis of the core wheel. This allows the wheels to move in two directions. Omni-wheeled robots can move at any angle in any direction, without rotating beforehand.

We use custom-designed platform with three 38-mm-diameter omni-directional wheels, driven by micro stepper motors to precisely control the rotation angle of each wheel. Having a holonomic system allow the robots to move smoothly and easily respond to user interaction.

\begin{figure}[htbp]
    \centering
    \includegraphics[width=\columnwidth]{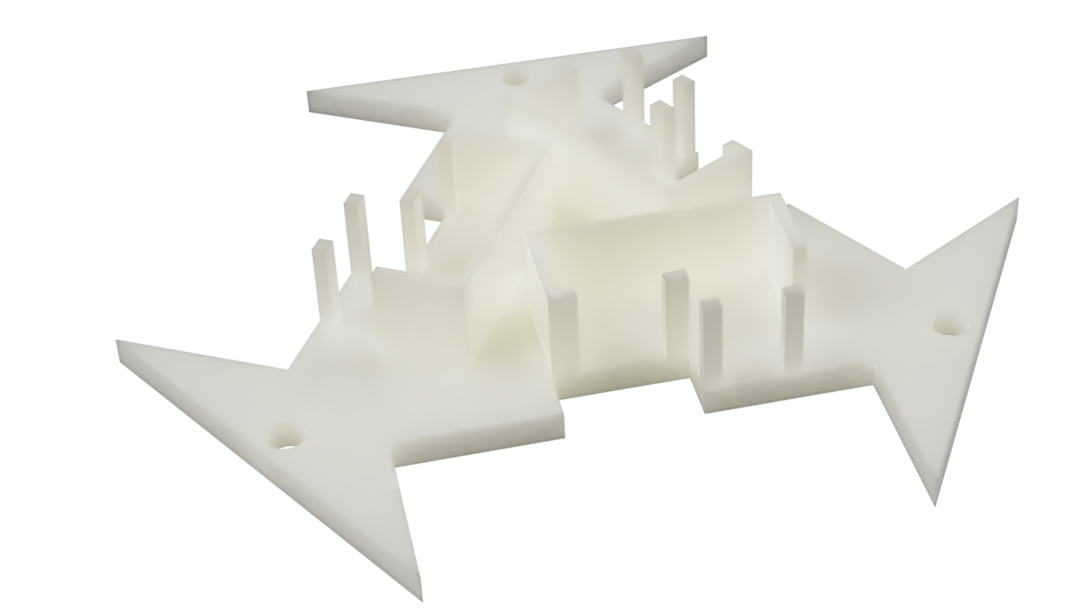}
    \caption{3D Model of Misaka}
    \label{fig:3DModel}
\end{figure}

Most of the components are 3D printed using PLA(Polylactic Acid), which allows for easy replication and modification. Currently, a Misaka takes about 3 hours to print with a JGAURORA A3S 3D printer using an infill of 20\%. We use Fusion 360 for CAD design and Ultimaker Cura for slicing. 3D Model is shown in Fig~\ref{fig:3DModel}.

To drive the robot, a motor driver chip (DRV8825) and three 2 phase 4 wire Stepper Gear Motor (CHS-GM1024-10BY) are used. With this combination, the robot has a maximum speed of approximately 20 cm/s.

8 capacitive touch buttons which independently back-illuminated in full RGB using WS2812B are wrapped inside the 3D printed enclosure to provide capacitive touch sensing capabilities and the robot’s state display as well as full color indicating.

\begin{figure}[htbp]
    \centering
    \includegraphics[width=0.6\columnwidth]{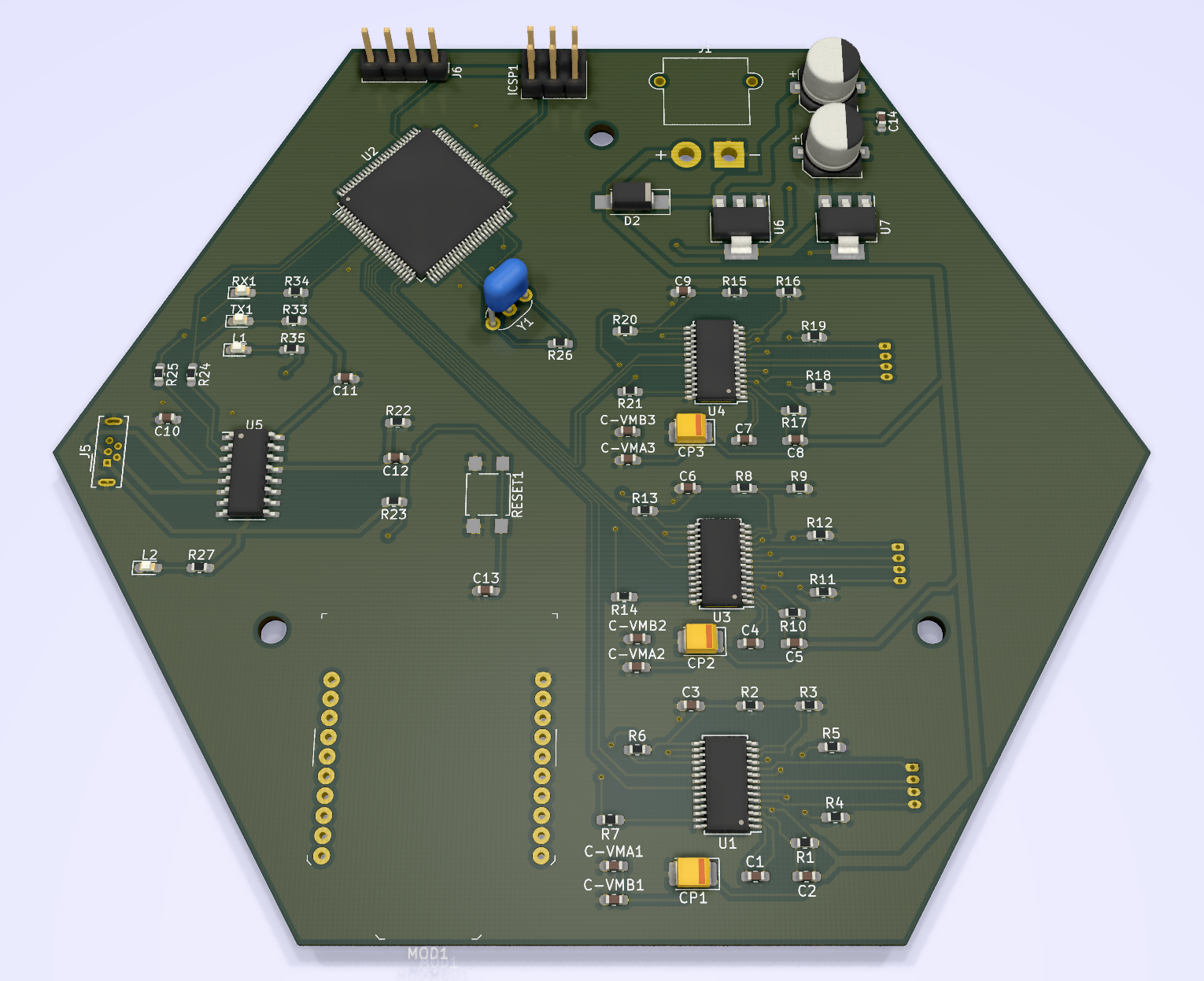}
    \caption{Top-down view of the main Misaka circuit board}
    \label{fig:PCB}
\end{figure}

The main Misaka circuit board shown in Fig~\ref{fig:PCB}, allows for robot control. The main processors onboard Misaka is a high-performance, low-power 16MHz 8-bit AVR microcontrollers(Microchip ATmega2560-16AU) that combines 256KB ISP flash memory, 8KB SRAM, 4KB EEPROM, 86 general purpose I/O lines, 32 general purpose working registers, real time counter, six flexible timer/counters with compare modes, PWM, 4 USARTs, byte oriented 2-wire serial interface, 16-channel 10-bit A/D converter, and a JTAG interface for on-chip debugging. ATmega2560 manages the overall logic computation.

Most of the power in the robots are consumed by the motors. The current draw of each robot is approximately 200 mA when the motors are stalled and 800 mA during typical use. Thus, with a 450 mAh battery, robots are capable of moving for half an hour, and can work even longer under normal usage.

Each robot communicates with each other using Digi XBee module. Mesh networking is used in our application to stimulate smart grid communication protocol such as DigiMesh and Zigbee3.0, which RF line of sight range can reach 2 miles.

Mesh networking is used in electric grid applications where the range between two points may be beyond the range of the two radios located at those points, but intermediate radios are in place that could forward on any messages to and from the desired radios.



A real-time high-accuracy 2D localization tracking system for mobile robots, based on simple printed patterns, building over algorithms previously developed for digital pens similar to Hostettler\cite{hostettler2016real} is used for Misaka position tracking. It need a microdot pattern printed paper under Misaka to work functionally. However, we can print other pattern on this specific paper without affecting any function. Also the pattern printed paper can be made translucent to be laid on the screen or other paper.

We use SONiX OID SNM9S500C3000A Decoder SoC Module, which is a highly integrated system-on-chip (SoC) module embedded CMOS Sensor and OID image decoder, to transmit location data to MCU.


Users can modify Misaka for their applications by designing custom modules that attach to its core module or adding powerful chips and development boards to achieve more functions, such as computer vision, wifi, machine learning algorithms, etc.

\begin{figure}[htbp]
    \centering
    \includegraphics[width=0.8\columnwidth]{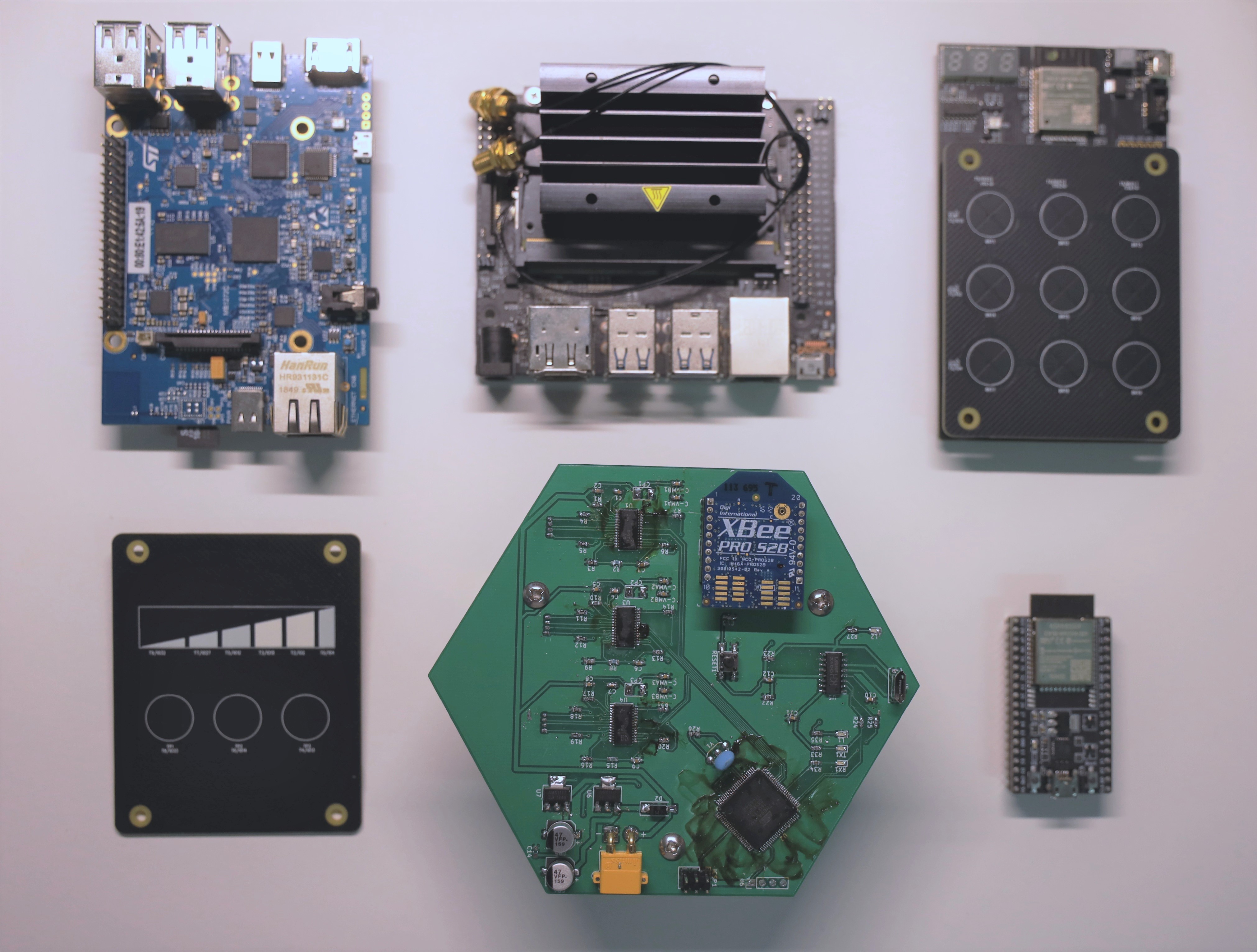}
    \caption{Misaka core and its compatible extensions}
    \label{fig:extension}
\end{figure}

Misaka core and its compatible extensions are shown in Fig~\ref{fig:extension}.

For example, in order to give Misaka Linux development environment and capabilities of testing machine learning algorithms, we add Nvidia Jetson Nano to Misaka core through UART for high-level control and image processing. And to connect them with bluetooth and wifi, we add ESP32 modules which also interact with Misaka through serial communication.







\section{Conclusion}

We introduced Misaka, a visualized swarm testbed for smart grid algorithm evaluation, also a platform for developing tabletop tangible swarm interfaces.

We propose a simplified, fully decentralized average consensus algorithm to solve DED problems and visualize it with Misakas in various ways.

We use some scenarios developed with Misaka to illustrate the potential of this user interface, including dynamic iteration process physicalization, information transmission visualization, time-series navigation and multiple scatterplots drawing.

We hope that this paper and Misaka platform will spur more research in smart grid distributed algorithm development and interactive testbed.

All necessary material and documentation for implementing Misaka can be found at

\url{https://github.com/TingliangZhang/Misaka}


\bibliographystyle{IEEEtran}
\bibliography{IEEEabrv,refs}

\end{document}